# Research on Predicting Public Opinion Event Heat Levels Based on Large Language Models

Yi Ren
Lab. of AI, Hangzhou Institute of Technology of Xidian University
Hangzhou, China
22171214767@stu.xidian.edu.cn

Tianyi Zhang
School of Artificial Intelligence, Xidian University
Xi'an, China
22171110624@stu.xidian.edu.cn

Weibin Li*
Lab. of AI, Hangzhou Institute of Technology of Xidian University
Hangzhou, China
weibinli@xidian.edu.cn

DuoMu Zhou
Lab. of AI, Hangzhou Institute of Technology of Xidian University
Hangzhou, China
zhouzdm@gmail.com

Chenhao Qin
School of Artificial Intelligence, Xidian University
Xi'an, China
22171214750@stu.xidian.edu.cn

FangCheng Dong
School of Information and Communication Engineering, North University of China
Taiyuan, China
dongf343@gmail.com

*Abstract*—In recent years, with the rapid development of large language models, serval models such as GPT-4o have demonstrated extraordinary capabilities, surpassing human performance in various language tasks. As a result, many researchers have begun exploring their potential applications in the field of public opinion analysis. This study proposes a novel large-language-models-based method for public opinion event heat level prediction. First, we preprocessed and classified 62,836 Chinese hot event data collected between July 2022 and December 2023. Then, based on each event's online dissemination heat index, we used the MiniBatchKMeans algorithm to automatically cluster the events and categorize them into four heat levels (ranging from low heat to very high heat). Next, we randomly selected 250 events from each heat level, totalling 1,000 events, to build the evaluation dataset. During the evaluation process, we employed various large language models to assess their accuracy in predicting event heat levels in two scenarios: without reference cases and with similar case references. The results showed that GPT-4o and DeepseekV2 performed the best in the latter case, achieving prediction accuracies of 41.4% and 41.5%, respectively. Although the overall prediction accuracy remains relatively low, it is worth noting that for low-heat (Level 1) events, the prediction accuracies of these two models reached 73.6% and 70.4%, respectively. Additionally, the prediction accuracy showed a downward trend from Level 1 to Level 4, which correlates with the uneven distribution of data across the heat levels in the actual dataset. This suggests that with the more robust dataset, public opinion event heat level prediction based on large language models will have significant research potential for the future.

*Keywords—large language model, Public Opinion Analysis, Event Heat Prediction, GPT-4o.*

## I. INTRODUCTION

In recent years, the emergence of large language models (LLMs), such as OpenAI's ChatGPT[1], has brought profound changes to the field of natural language processing (NLP). With their powerful few-shot[2] and zero-shot learning capabilities, these models exhibit remarkable generalization performance, enabling them to handle various complex language tasks and generate coherent, logically consistent responses. As a result, LLMs have demonstrated broad potential applications across multiple fields.

Although ChatGPT remains closed-source, the development of the open-source community has provided researchers with more alternatives. With the introduction of a series of high-performance open-source models like LLaMA[3], Qwen[4], and ChatGLM[5], researchers now have more opportunities to explore the application of LLMs in different domains, achieving notable results. For example, in the medical field[6], Wang et al. proposed the BenTsao[7] model, which integrates medical knowledge graphs and literature and utilizes a Chinese medical instruction tuning dataset generated via the ChatGPT API to fine-tune models like LLaMA. In the legal field, Zhou et al. developed LawGPT[8], which underwent secondary pretraining and instruction tuning on a large-scale Chinese legal corpus, endowing it with robust legal question-answering capabilities. In the field of remote sensing[9, 10], Kuckreja et al. proposed GeoChat, the first multimodal large model[11] capable of understanding various types of remote sensing images. In the hydrology field, Ren et al. introduced WaterGPT[12], which, based on Qwen-7B-Chat and Qwen2-7B-Chat, underwent large-scale secondary pretraining and instruction tuning on domain-specific data, enabling professional knowledge Q&A and intelligent tool usage. In the field of sentiment analysis, Peña et al. experimentally evaluated the performance of four Spanish LLMs in classifying public affairs documents, demonstrating that LLMs can effectively handle and understand complex language documents, classifying them into up to 30 topics, thus providing technical support for promoting transparency, accountability, and civic participation[13].

Despite extensive research exploring the application of LLMs in specialized domains, studies focused on predicting the influence of trending events remain limited. We propose a public opinion event heat level prediction method based on LLMs to address this. In this study, we first preprocessed and classified 62,836 data points covering trending events in China from July 2022 to December 2023. Based on each event's network dissemination heat index, we used the MiniBatchKMeans algorithm to perform automated clustering, categorizing these events into four heat levels (from Level 1 to Level 4). Subsequently, we randomly selected 250 events from each heat level, totaling 1,000 events, as the evaluation dataset. During the evaluation process, we used various LLMs to assess their accuracy in predicting the heat level of events under two scenarios: with and without reference cases.

The results showed that GPT-4o[14] and Deepseekv2[15] performed the best, achieving prediction accuracies of 41.4% and 41.5%, respectively, in scenarios with similar case references. Although the overall prediction accuracy was still low for low-heat (Level 1) events, the prediction accuracies of these two models reached 73.6% and 70.4%, respectively. Additionally, the prediction accuracy decreased from Level 1 to Level 4, related to the uneven distribution of data across heat levels in the actual dataset. This suggests that with further expansion of the dataset, LLM-based public opinion event heat prediction holds promising research potential.

The main contributions of this paper are as follows:

1. A novel LLM-based method for predicting public opinion event heat levels is proposed.

2. A comprehensive evaluation of the performance of state-of-the-art LLMs in public opinion event heat level prediction and evaluation results under different scenarios (such as zero-shot and few-shot) is provided.

## II. PROPOSED METHODS

### A. Overall architecture

In this study, we propose a method for predicting the heat level of public opinion events based on large models. This method consists of three main modules: data processing, public opinion event heat level classification, and model prediction. The detailed process is illustrated in Figure 1.

First, in the data processing module, we organize and refine the content of each public opinion event and categorize them based on their attributes. Then, in the public opinion event heat level classification module, we use a 62,836 public opinion events dataset and apply the MiniBatchKMeans algorithm to cluster them based on their online dissemination heat index automatically. Through this process, we classify public opinion events into four heat levels: low, medium, high, and very high.

Finally, in the model prediction module, we use the categorized dataset from the data processing module to train the bge-large-zh-1.5[16] embedding model. When utilizing the evaluation dataset, the model recalls ten similar public opinion events based on the content of the given event using the bge-large-zh-1.5 model. Subsequently, we fill the content of these similar events into a predefined template and input it into a large language model. The model then generates the final predicted heat level of the public opinion event, which is recorded for further analysis.

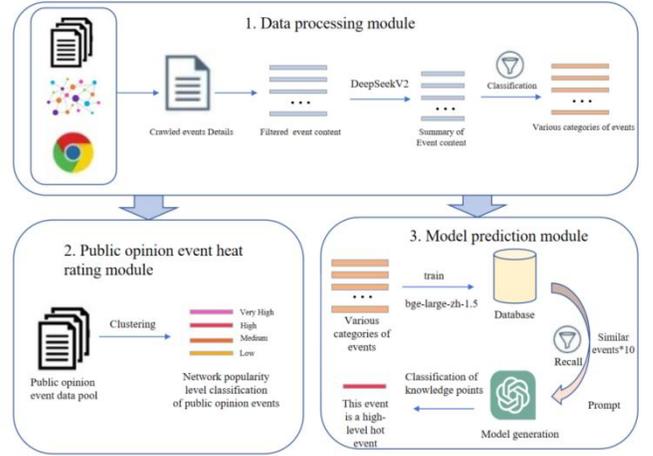

Fig. 1. Various tasks that our remote sensing multi-modal large model can complete

### B. Data processing module

Our original dataset contains 62,836 records covering hot events in China from July 2022 to December 2023. In the data processing module, we first crawled detailed information for each public opinion event based on its title from the internet. After filtering out events with empty content or garbled text, we retrieved detailed information for 40,081 public opinion events.

Next, we used the DeepSeekV2 API to extract summaries from the detailed content of each event, condensing it into a concise description to serve as the content representation of the event. For events where detailed information could not be retrieved, we used the event's title as its content description.

We manually categorized each event into one of 20 categories, including transportation, sports, agriculture, healthcare, and others. We then constructed a dataset for training the embedding model. To balance the data distribution, the number of events in each category was capped at 3,000, with excess entries being discarded. The entries in the dataset were used to create positive and negative samples based on the main content of the events: entries in the same category were treated as positive samples, while entries in different categories were treated as negative samples. Each training sample consisted of the event's content, content from another event in the same category (positive sample), and content from an event in a different category (negative sample). Ultimately, the training dataset contained 33,864 records, with the specific distribution and proportion of categories shown in Figure 2.

Identify applicable funding agency here. If none, delete this text box.

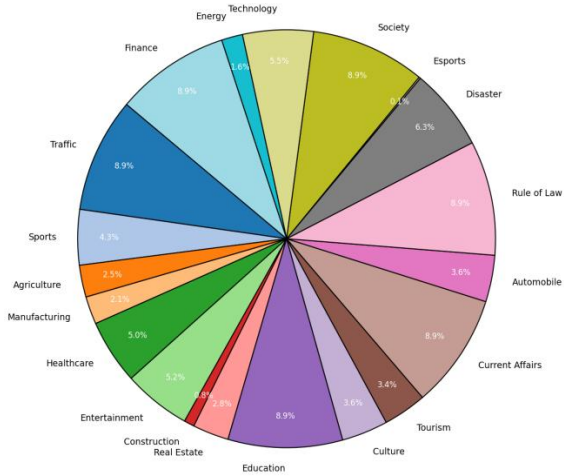

Fig. 2. Various types and proportions of embedding model training data sets

## C. Public opinion event heat rating module

In this module, we applied the MiniBatchKMeans algorithm for automated clustering based on the online propagation heat index of public opinion events. The events were categorized into four heat levels: low,, medium, high, and very high(ranging from level one to level four). We randomly selected 250 events from the public opinion event pool from each heat level, resulting in a total of 1,000 events to construct an evaluation dataset for large language models.

The clustering process is as follows:

（1）Calculation of the Sum of Squared Errors (SSE):

$$SSE = \sum_{j=1}^{P} \sum_{V_i \in C_j} \| V_i - \mu_j \|^2 \quad (1)$$

Here, P represents the number of clusters, $C_j$ denotes the *jth* cluster, and $\mu_j$ is the centroid of the *jth* cluster. $V_i$ represents the vector belonging to the jth cluster. The SSE measures the sum of the distances between data points and their respective cluster centroids, serving as one of the indicators to evaluate clustering performance. A smaller SSE indicates that the points within a cluster are more tightly grouped. By plotting the SSE values for different numbers of clusters P, one can preliminarily assess the reasonable range for the number of clusters.

（2）Calculation of the Silhouette Coefficient:

$$s(i) = \frac{b(i) - a(i)}{\max(a(i), b(i))} \quad (2)$$

Here, $a(i)$ represents the average distance from data point $i$ to all other points within the same cluster, and $b(i)$ represents the average distance from data point i to the nearest points in a different cluster. The silhouette coefficient $S$ for the entire dataset is the average of the silhouette scores s(i) for all data points:

$$S = \frac{1}{N} \sum_{i=1}^{N} s(i) \quad (3)$$

Here, *N* represents the total number of data points.

（3）Selection of the optimal number of clusters:

$$p = \arg\max_{k} S(k) \quad (4)$$

Here, *S(k)* represents the silhouette coefficient for different numbers of clusters *k*, and *p* is the optimal number of clusters that maximizes *S(k)*.

After determining the optimal number of clusters, denoted as p, the K-Means algorithm is applied for clustering. The update formula for the cluster centers is as follows:

$$C_j = \frac{1}{|S_j|} \sum_{x_i \in S_i} x_i \quad (5)$$

where $C_j$ is the *j*-th cluster center, $S_j$ represents the set of all data points belonging to the *j*-th cluster, and $X_i$ is the *i*-th data point.

The final clustering results are shown in Table 1.

TABLE I
DATA DISTRIBUTION AFTER CLUSTERING.

| Heat level | Internet communication popularity index range | Number of events |
|---|---|---|
| Low heat level | [0.000000,8.777964) | 54789 |
| Medium heat level | [8.777964,21.462457) | 5719 |
| High heat level | [21.462457,42.399911) | 2000 |
| Very high heat level | [42.399911 ,Inf) | 328 |

## D. Model prediction module

In this module, we will use the training dataset generated in the previous data processing stage to train the bge-large-zh-1.5 model. After training is completed, the model will retrieve similar events for each public opinion event in the evaluation dataset. Specifically, the model will recall 10 similar public opinion events for each input event and output their content information, online heat propagation index, and heat level.

Based on the retrieval results, we will construct a template as shown in Appendix, and the content of this template will be input into the model.

We employed six of the most advanced large language models currently available on the market, including API-based and locally deployed models. Details are provided in Table 2.

TABLE II
THE MODEL EVALUATED IN THIS ARTICLE.

| Model | Creator | #Parameters | Access |
|---|---|---|---|
| GPT-4o | OpenAI | undisclosed | API |
| DeepSeek-V2 | DeepSeek-AI | 236B | API |
| GLM-4 | Zhipu AI | undisclosed | API |
| GLM-4-9B-chat | Zhipu AI | 9B | Weights |
| Qwen2-7B-instruct | Alibaba Group | 7B | Weights |
| InternLM2.5-7B-chat | Shanghai AI Lab | 7B | Weights |

## III. EXPERIMENTS AND ANALYSIS

### A. Experimental setup

We evaluated the ability of different large language models to predict the heat levels of public opinion events, primarily based on two approaches: direct prediction without any case references and prediction after referencing ten similar event cases. Our specific template is shown in the Appendix. We employed a multiple-choice format, allowing the language model to choose one of the four heat levels. This method helps standardize the model's output.

Our evaluation results are divided into overall prediction accuracy and the level-specific prediction accuracy for 250 events in each heat level category.

### B. Embedding Model Training Setup

To achieve similar event recall, we trained the bge-large-zh-1.5 model using the training dataset introduced in Section 2.1. The model was trained for one epoch, and the trained model was mixed with the original model in a 1:1 ratio. This approach balances specialized and general capabilities. The specific training parameters are shown in Table 3.

TABLE III
TRAIN PARAMETERS.

| Hyper parameter | Value |
| --- | --- |
| Precision | fp16 |
| Epochs | 1 |
| Per_device_train_batch_size | 18 |
| Query_max_len | 256 |
| Passage_max_len | 256 |
| Learning rate | $5e^{-5}$ |

To further evaluate the performance of the trained embedding model, we used the evaluation dataset constructed in Section 2.2. First, we recalled ten similar events for each public opinion event, and the heat level of the event was determined by the most frequent heat level among the ten recalled similar events. The final result is shown as Scenario 1 in Table 4. Additionally, to further observe the distribution of heat levels among the recalled events, we chose the most frequent and the second most frequent heat levels from similar events to determine the heat level of the event. If either result was correct, the event's heat level prediction was considered correct. This result is shown as Scenario 2 in Table 4.

TABLE IV
PERFORMANCE EVALUATION OF EMBEDDING MODEL AFTER TRAINING.

| Heat level | Scenario 1 | Scenario 2 |
| --- | --- | --- |
| Low | 100.00 | 100.00 |
| Medium | 0.00 | 87.60 |
| High | 1.20 | 17.60 |
| Very high | 0.00 | 3.60 |

As shown in the table, in the case of low heat events, both Scenario 1 and Scenario 2 achieved 100.00% prediction accuracy. For medium heat events, Scenario 1 had 0% accuracy, while Scenario 2 maintained a relatively high accuracy of 87.60%. Regarding high heat events, Scenario 1 and Scenario 2 achieved lower accuracies, with 1.20% and 17.60%, respectively. For very high heat events, both Scenario 1 and Scenario 2 had even lower accuracies, at 0% and 3.60%, respectively. From these results, it can be observed that as the heat level increases, the prediction accuracy decreases progressively. This is primarily due to the uneven distribution of the real dataset, where the number of events from medium to very high heat levels combined is less than one-fifth of the number of low-heat events.

### C. Large language model prediction results

To comprehensively evaluate the performance of large language models in predicting the heat levels of public opinion events, we carefully designed two scenarios: with and without case references. In the "without case references" scenario, the model predicts the heat level based on the event content. The "with case references" scenario has two setups: one with actual similar events and one with simulated cases. In the actual case setup, we used the trained embedding model to recall ten similar events for each event in the evaluation dataset, which were then integrated into a designed prompt and fed into the model for prediction. In the simulated case setup, due to the uneven distribution of the real dataset, we randomly selected three events from each of the heat levels above the medium level (for a total of nine events), incorporated them into a designed prompt, and fed it into the model for prediction. This approach simulates the model's prediction results when the dataset is more balanced. The experimental results are shown in Table 5 and Figure 3.

TABLE V
PREDICTION RESULTS OF PUBLIC OPINION EVENT POPULARITY BY VARIOUS LARGE LANGUAGE MODELS IN DIFFERENT SCENARIOS.

| Model | without case references | with case references | with case references(simulated situation) |
| --- | --- | --- | --- |
| GPT-4o | 28.10 | 30.30 | 41.40 |
| DeepSeek-V2 | 24.40 | 30.30 | 41.50 |
| GLM-4 | 23.77 | 27.57 | 41.34 |
| GLM-4-9B-chat | 25.20 | 23.30 | 26.20 |
| Qwen2-7B-instruct | 23.80 | 27.70 | 32.70 |
| InternLM2.5-7B-chat | 25.00 | 25.00 | 24.10 |

As shown in Table 5, the results of the large language models in the direct prediction scenario were generally poor. The highest accuracy, for instance, was 28.10% from GPT-4o. The prediction results of larger models accessed via APIs were not significantly different from locally run models. For example, DeepSeek-V2 and GLM4 achieved accuracies of 24.40% and 23.77%, respectively, while the best-performing local model, GLM-4-9B-chat, reached 25.20%.

It is worth noting that, as shown in Figure 3, the level-specific prediction results show that GLM4 achieved 70.0% accuracy for high-heat events, Qwen2-7B-instruct reached 65.6% accuracy for very high heat events, and GLM-4-9B-chat had 56.8% accuracy for medium-heat events.

In the "with case references" scenario, all models except GLM-4-9B-chat and InternLM2.5-7B-chat showed improvements in prediction accuracy. The best-performing models, GPT-4o and DeepSeek-V2, both achieved a prediction accuracy of 30.3%. As shown in Figure 3, the prediction accuracy of GPT-4o and DeepSeek-V2 decreased as the heat level increased. For low-heat events, both models achieved 73.6% and 70.4% accuracy, respectively. However, the accuracy progressively declined for medium and higher heat levels, which aligns with the uneven distribution of the real dataset. The poorer prediction performance for events at the medium and higher heat levels is due to the uneven data distribution and missing data.

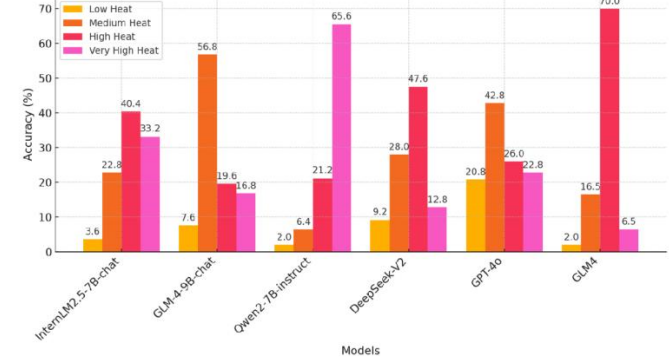

(a) There are no similar cases

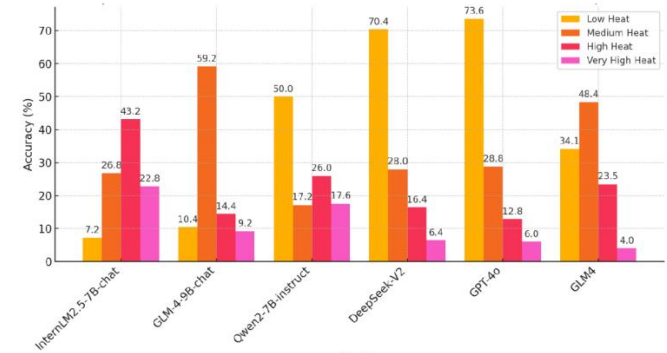

(b) There are similar cases

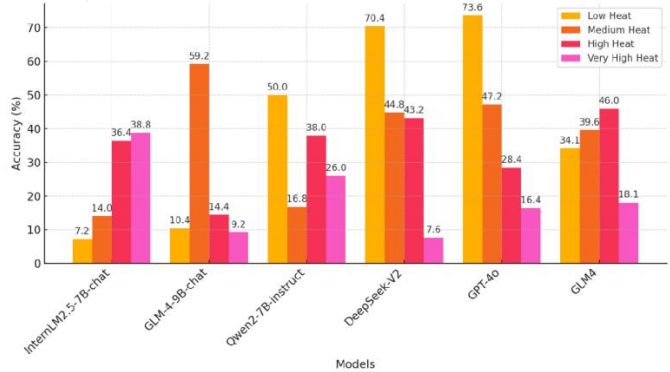

(c) There are similar cases (simulated situations)

Fig. 3. Large language models predict results for each level of events in different scenarios.

In the simulated case scenario, all models except InternLM2.5-7B-chat showed further improvements in accuracy. GPT-4o and DeepSeek-V2 achieved optimal prediction accuracies of 41.40% and 41.50%, respectively. As shown in Figure 3, the prediction accuracy for both models still followed a decreasing trend as heat levels increased. For DeepSeek-V2, the accuracy was 44.80% for medium-heat events, 43.20% for high-heat events, and 7.6% for very high-heat events. GPT-4o achieved 47.20%, 28.4%, and 16.4% for medium, high, and ultra-high heat levels. These results suggest that the prediction accuracy of the models is influenced by the quality of the similar events provided. Due to the random selection process, it was difficult to match the most similar events for reference, which affected the model's prediction accuracy. Additionally, large language models tend to be conservative when predicting.Without a clear reference to similar very high heat events, the models tend to predict lower heat levels for very high heat events.

Furthermore, although the GLM4 model achieved an overall accuracy of 41.40%, its prediction results did not follow a smooth decreasing trend. Instead, it achieved accuracies of 34.1%, 39.6%, and 46.0% for low, medium, and high heat levels, respectively. We believe this is because the model, when referencing similar events, tends to focus on a few specific events rather than considering the whole set, leading to prediction bias.

IV. CONCLUSION

This study evaluated the performance of various large language models in predicting the heat levels of public opinion events, focusing on analyzing changes in prediction accuracy with and without reference to similar cases. The experimental results show that the direct prediction performance of large language models without reference cases was relatively poor, with the best model, GPT-4o, achieving only 28.10% accuracy. Nevertheless, certain models performed better in specific heat level categories. For instance, GLM4 reached 70.0% accuracy for high-heat events, and Qwen2-7B-instruct achieved 65.6% accuracy for very high heat events.

When similar case references were available, the overall prediction performance of the models improved, with both GPT-4o and DeepSeek-V2 achieving 30.30% accuracy. In the simulated case scenario, especially, the prediction accuracy of GPT-4o and DeepSeek-V2 reached 41.40% and 41.50%, respectively. However, it is important to note that the prediction accuracy of the models decreased as the heat level increased, particularly for events at medium heat levels and above. This decline is closely related to the uneven distribution of the dataset, where the lack of sufficient samples at the higher heat levels resulted in poorer prediction performance. For low-heat events, GPT-4o and DeepSeek-V2 performed exceptionally well, achieving prediction accuracies of 73.6% and 70.4%, respectively. In contrast, for very high-heat events, the models generally performed poorly, which may be due to the insufficient quality of similar cases and the models' tendency towards conservative predictions.

Overall, although large language models still face challenges in predicting the heat levels of public opinion events, such as uneven data distribution and difficulties in matching similar cases, their strong performance in predicting low-heat events and the overall improvement in prediction accuracy suggest that public opinion analysis based on large language models has significant research potential. Future research could improve prediction accuracy for events at different heat levels by optimizing dataset distribution and enhancing the mechanism for matching similar cases.

*APPENDIX*

**No case prompt:**

The event content is {event}### Please predict the popularity level that the event will reach based on the above event content. Please output the options. Please select only the most relevant level. \n{options}

**There are similar cases prompt:**

The event content is {event}### Please predict the popularity level that the event will reach based on the above event content. Please output the options. Please select only the most relevant level. \n{options}Refer to similar event information as {Case}.

Options="""Option: A, heat level 1, heat index range is (0.000000,8.777964)

Option: B, heat level 2, heat index range is (8.777964,21.462457)

Option: C, heat level 3, heat index range is (21.462457,42.399911)

Option: D, heat level 4, heat index range is (42.399911,Inf)"""

REFERENCES

[1] A. Bahrini, M. Khamoshifar, H. Abbasimehr, et al., "ChatGPT: Applications, opportunities, and threats," in Systems and Information Engineering Design Symposium, vol. 274, 2023. *(references)*

[2] T. B. Brown, "Language models are few-shot learners," arXiv preprint arXiv:2005.14165, 2020.

[3] H. Touvron, T. Lavril, G. Izacard, et al., "Llama: Open and efficient foundation language models," arXiv preprint arXiv:2302.13971, 2023.

[4] H. Touvron, T. Lavril, G. Izacard, et al., "Llama: Open and efficient foundation language models," arXiv preprint arXiv:2302.13971, 2023.

[5] Z. Du, Y. Qian, X. Liu, et al., "GLM: General language model pretraining with autoregressive blank infilling," arXiv preprint arXiv:2103.10360, 2021.

[6] Y. Ren, W. Li, L. Shi, J. Ding, J. Du, and T. Chen, "FUO_ED: A dataset for evaluating the performance of large language models in diagnosing complex cases of fever of unknown origin," SSRN, no. 4952379, 2024.

[7] H. Wang, C. Liu, N. Xi, et al., "Huatuo: Tuning llama model with Chinese medical knowledge," arXiv preprint arXiv:2304.06975, 2023.

[8] Z. Zhou, J.-X. Shi, P.-X. Song, et al., "LawGPT: A Chinese legal knowledge-enhanced large language model," arXiv preprint arXiv:2406.04614, 2024.

[9] C. H. Qin, W. B. Li, T. Y. Zhang, et al., "Improved DeepLabv3+ based flood water body extraction model for SAR imagery," in IGARSS 2024-2024 IEEE International Geoscience and Remote Sensing Symposium, pp. 1196-1199, 2024.

[10] T. Zhang, W. Li, X. Feng, et al., "Super-resolution water body extraction based on MF-SegFormer," in IGARSS 2024-2024 IEEE International Geoscience and Remote Sensing Symposium, pp. 9848-9852, 2024.

[11] Y. Ren, T. Zhang, Z. Han, et al., " A Novel Adaptive Fine-Tuning Algorithm for Multimodal Models: Self-Optimizing Classification and Selection of High-Quality Datasets in Remote Sensing,'' arXiv preprint arXiv:2409.13345, 2024.

[12] Y. Ren, T. Zhang, X. Dong, et al., "WaterGPT: Training a large language model to become a hydrology expert," SSRN, no. 4863665, 2024.

[13] A. Peña, A. Morales, J. Fierrez, et al., "Leveraging large language models for topic classification in the domain of public affairs," in International Conference on Document Analysis and Recognition, Springer Nature Switzerland, pp. 20-33, 2023.

[14] J. Achiam, S. Adler, S. Agarwal, et al., "GPT-4 Technical Report," arXiv preprint arXiv:2303.08774, 2023.

[15] A. Liu, B. Feng, B. Wang, et al., "Deepseek-v2: A strong, economical, and efficient mixture-of-experts language model," arXiv preprint arXiv:2405.04434, 2024.

[16] P. Zhang, S. Xiao, Z. Liu, et al., "Retrieve anything to augment large language models," arXiv preprint arXiv:2310.07554, 2023.